\documentclass[conference]{IEEEtran}
\IEEEoverridecommandlockouts
\usepackage{graphics} 
\usepackage{epsfig} 
\usepackage{mathptmx} 
\usepackage{times} 
\usepackage{amsmath} 
\usepackage{amssymb}  
\usepackage{comment} 
\usepackage{hyperref}
\usepackage{color}
\usepackage{stfloats}
\usepackage{algorithmic}
\usepackage[linesnumbered,ruled,vlined]{algorithm2e}
\usepackage{mathtools}
\usepackage{adjustbox}
\usepackage{booktabs}
\usepackage{blindtext}
\usepackage{verbatim}
\usepackage{xspace}
\usepackage{newtxtext,newtxmath}
\usepackage{array,multirow,graphicx}
\usepackage{subcaption}
\usepackage{booktabs}
\usepackage{bm}
\usepackage[dvipsnames]{xcolor}

\begin{document}

\title{Divide-Conquer Transformer Learning for Predicting Electric Vehicle Charging Events \\Using Smart Meter Data
\thanks{This work was supported in part by the Australian Research Council (ARC) Discovery Early Career Researcher Award (DECRA) under Grant DE230100046.}
}

\author{\IEEEauthorblockN{Fucai Ke\textsuperscript{1}, Hao Wang\textsuperscript{2,3,*}}
\IEEEauthorblockA{
\textsuperscript{1}Department of Human Centred Computing, Faculty of Information Technology, Monash University, Australia \\
\textsuperscript{2}Department of Data Science and AI, Faculty of Information Technology, Monash University, Australia \\
\textsuperscript{3}Monash Energy Institute, Monash University, Australia
}
\thanks{*Corresponding author: Hao Wang (hao.wang2@monash.edu).}
}

\maketitle

\begin{abstract}
Predicting electric vehicle (EV) charging events is crucial for load scheduling and energy management, promoting seamless transportation electrification and decarbonization. 
While prior studies have focused on EV charging demand prediction, primarily for public charging stations using historical charging data, home charging prediction is equally essential. However, existing prediction methods may not be suitable due to the unavailability of or limited access to home charging data. 
To address this research gap, inspired by the concept of non-intrusive load monitoring (NILM), we develop a home charging prediction method using historical smart meter data. Different from NILM detecting EV charging that has already occurred, our method provides predictive information of future EV charging occurrences, thus enhancing its utility for charging management.
Specifically, our method, leverages a self-attention mechanism-based transformer model, employing a ``divide-conquer'' strategy, to process historical meter data to effectively and learn EV charging representation for charging occurrence prediction. Our method enables prediction at one-minute interval hour-ahead. Experimental results demonstrate the effectiveness of our method, achieving consistently high accuracy of over 96.81\% across different prediction time spans. Notably, our method achieves high prediction performance solely using smart meter data, making it a practical and suitable solution for grid operators.

\end{abstract}


\section{Introduction}\label{sec:intro}
As the popularity of electric vehicles (EVs) continues to surge, EV charging demand is increasing significantly. 
Uncoordinated EV charging poses substantial pressure on the power grid, primarily due to the considerable increase in peak load on distribution networks \cite{luo2011forecasting, islam2016feasibility}. 
The lifespan of network assets can be significantly impacted by overloading, as a result of increasing EV charging demand at homes \cite{Muratori2018, hoque2022network, liu2021planning}.
The growing adoption of EVs presents new challenges for distribution system operators to plan and operate their systems. Therefore, it is essential to advance predictive models for EV charging events or called activities. Such predictive EV charging information is crucial for ensuring the stability and reliability of the network and holds substantial value for system operators \cite{frendo2020improving, cui2023stacking}.

In recent years, there has been a large body of literature on various methods for predicting EV charging demand and enhancing EV charging management \cite{qian2010load}. 
Such prediction studies often relied on a wide range of factors, including traffic data, ride-hailing trip data, arrival time, and charging rates \cite{xing2019charging, arias2016electric, li2012modeling, wang2016two, yu2015centralized}, and utilized various optimization algorithms or machine learning models to predict different charging variables like charging rate \cite{nie2016system}, charging duration \cite{beaude2016reducing} and charging power \cite{tang2016aggregated}. 
All these methods provide useful insights into EV charging demand, primarily at charging stations \cite{you2015optimal}. However, a substantial number of charging events/activities occur at home rather than at charging stations, making the above existing studies less evident to system operators. 
More importantly, the above studies leveraged historical EV charging data to make predictions. However, behind-the-meter at-home charging is highly invisible to the operators and home charging data may not be available, making home charging event prediction even more challenging.

Several recent non-intrusive methods have gained prominence in various domains, for example, enhancing grid management through EV charging detection and disaggregation \cite{wang2022evsense, wang2020deep, Zhang2014, zhou2020non, Martin2023nonintrusive}. 
These innovative studies have demonstrated commendable performance in post-charging analysis or real-time detection, but a critical aspect remains unexplored: predicting future EV charging events, particularly in the context of at-home charging. In contrast to post-charging behavioral analysis, predictive information of future EV charging events plays a vital role in energy management.
Therefore, \emph{our work aims to fill this void by developing an effective algorithm for home EV charging event prediction using solely historical smart meter data.}

This paper introduces a novel method, namely Divide-Conquer Transformer EV charging event prediction (DCT-EV), for predicting EV charging events several minutes ahead with high accuracy based on smart meter data. We evaluate our proposed method using a real-world dataset and compare it with four representative machine learning and deep learning models for different prediction requirements (minutes ahead). The main contributions of this paper are summarized as follow.
\begin{itemize}
    \item \textbf{New practical task:} Unlike existing studies on predicting EV charging at charging stations based on historical charging data, our work focuses on a more challenging and practical task of predicting home charging events using historical residential meter data. Our work builds upon the NILM concept to provide predictive information about future EV charging events, deviating from post-charging analysis.
    \item \textbf{Non-intrusive, minute-interval, high-accuracy model:} DCT-EV, a self-attention-based model with a transformer backbone, divides long historical smart meter time-series into multiple sub-sequences and processes them in parallel. This approach enables effective handling of meter data to serve EV charging event prediction. 
    \item \textbf{Extensive experiments and insights into EV charging event prediction using real-world datasets:} Our method, DCT-EV, outperforms benchmark models, establishing a new state-of-the-art in providing both 10-minute and 60-minute ahead EV charging event prediction. Extensive experiments provide valuable insights into 1) the impact of input history length on prediction performance, 2) the trade-off between model performance and the size of predictive time span, and 3) probability threshold selection.
\end{itemize}

The remainder of the paper is organized as follows. Section \ref{sec:method} presents the design of our DCT-EV model. Section \ref{sec:results} presents the experimental results, including comparisons between DCT-EV and benchmark models on a real-world dataset and the discussion of the evaluation results. Section \ref{sec:conclu} concludes the paper and outlines some future work directions.

\section{Methodology} \label{sec:method}
\begin{figure}[t]
    \centering
    \includegraphics[width=0.9\linewidth]{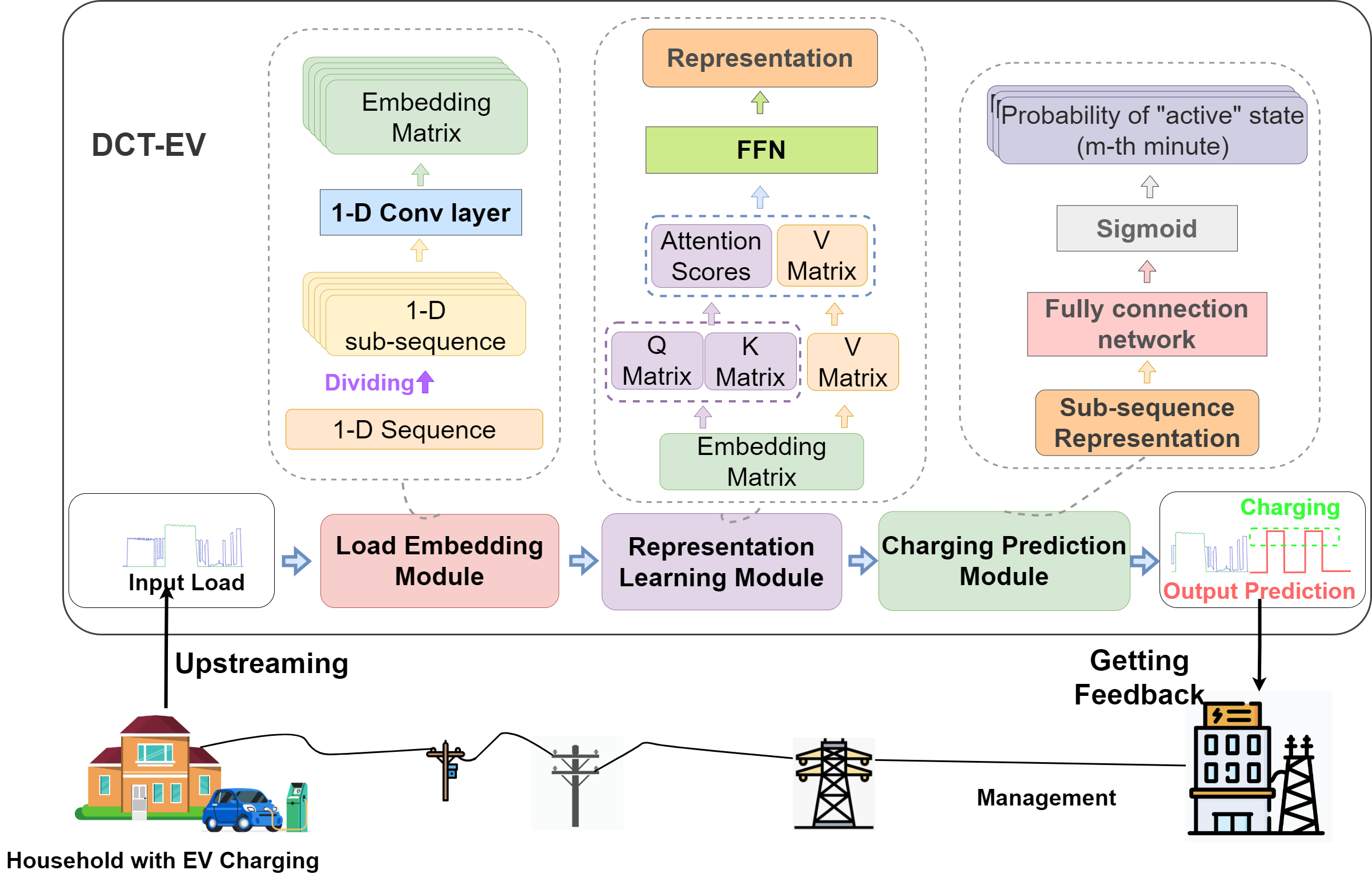}
    \caption{DCT-EV structure and application.}
    \label{dct_model_plot}
\end{figure}

The high-accuracy prediction of at-home EV charging events holds substantial value for energy management and grid operations. In this section, we present the structure of our proposed DCT-EV and its working principles in predicting future charging events using smart meter data, more specifically household load data. The model consists of three main components: the load embedding module, the EV representation learning module, and charging event prediction module, as illustrated in Figure~\ref{dct_model_plot}. 

In the electricity load embedding module, we first divide the lengthy smart meter data sequence into a few overlapping short sequences and then project the human-understandable and observable signals into high-dimension embedding vectors using convolutional layers. This approach can substantially reduce the computational resource and mitigate the inherent inductive bias for modeling load patterns. In the EV representation learning module, we boost the model's capacity by incorporating a self-attention-based transformer encoder. The encoder generates representations by taking into account the information from all sub-sequences. The charging event prediction module then predicts the near-future EV charging event and forecasts the probability of active EV charging over upcoming time steps.

\subsection{Load Embedding Module}
The objective of the load embedding module is to partition the long historical load sequence into several overlapping short sequences, drawing inspiration from \cite{he2023dynamic, he2021pruning, nie2022time}. This module can reduce computational resource consumption and enable the subsequent module to process the entire sequence in parallel.

The load embedding module takes each time series of electricity load from smart meter data as an input sequence, denoted as $\bm{x} = {x_{1}, ..., x_{t}, ..., x_{T}} \in \mathbb{R}^{1 \times T}$, where $T$ is the length of the input sequence. The long input sequence $\bm{x}$ is divided into $N = \frac{T-L}{stride} + 1$ number of overlapped sub-sequences, denoted as $\Bar{\bm{x}} = \{\bm{x}^1, ..., \bm{x}^n, ..., \bm{x}^N\} \in \mathbb{R}^{N \times L}$, based on sub-sequence length $L$ and a stride size of $stride$, where $\bm{x}^n = \{x^n_{1}, ..., x^n_{l},...x^n_{L}\} \in \mathbb{R}^{1 \times L}$. 

Each sub-sequence $\bm{x}^n$ is then projected to a high-dimensional latent space $\bm{\chi}^n = \{\chi^n_1,..., \chi^n_d,...,\chi^n_D\} \in \mathbb{R}^{1 \times D}$ using a convolutional layer as follows
\begin{equation}\label{con_eq}
\begin{split}
\bm{\chi}^n = \bm{x}^n\mathbf{\kappa},
\end{split}   
\end{equation}
where $D$ represents the dimension of the latent space, and $\mathbf{\kappa} \in \mathbb{R}^{L \times D}$ is the convolution layer kernal matrix.

The utilization of sub-sequence partitioning reduces information redundancy by treating each sub-sequence as a token instead of a single electricity load, thereby reducing the count of input tokens from $T$ to $N$. The subsequent component, the EV representation learning module and charging event prediction module, constitute a self-attention backbone, demanding substantial memory resources, depending upon the number of the input tokens. When calculating the attention scores, the computational resources required for these two different sequence sizes exhibit a notable difference, with approximately $T^2 >>  N^2$, showing the advantage of our sub-sequence partitioning. 

\subsection{EV Representation Learning Module}
The backbone of EV representation learning module is the self-attention mechanism, which is also a key element of the Transformer architecture \cite{vaswani2017attention}. We harness the attention mechanism to facilitate representation learning. Subsequently, we employ a lightweight linear layer to interpret the essential information and make predictions.

The $N$ embeddings of in-parallel sub-sequence are distinguished by their temporal order of interactions. To capture the intrinsic temporal dependencies within time series data, we implement \textbf{positional encoding}, a technique employing fixed sinusoids of varying frequencies \cite{vaswani2017attention}, to model the relative ordering of time series data as
\begin{equation}\label{pos_eq}
\begin{split}
\bm{\chi}_\text{pos} := \bm{\chi} + \bm{b}^\text{pos}.
\end{split}   
\end{equation}

\textbf{Multi-head attention layer} serves as a representation learner for EV charging behavior by computing the attention scores for all contextualized embeddings of each sub-sequence. Each sub-sequence is mapped into the queries, keys, and values in each head $h$. More specifically, the mapping can be expressed as
\begin{equation}\label{qkv_eq}
\begin{split}
&\boldsymbol{Q}_{h}= \bm{\chi}_\text{pos}\boldsymbol{W}_{h,Q} ,~\boldsymbol{K}_h= \bm{\chi}_\text{pos}\boldsymbol{W}_{h,K} ,~\boldsymbol{V}_h= \bm{\chi}_\text{pos}\boldsymbol{W}_{h,V} ,
\end{split}   
\end{equation}
where $\boldsymbol{W}_{Q}, \boldsymbol{W}_{K}, \boldsymbol{W}_{V} \in \mathbb{R}^{D\times D_m}$ are learnable mapping matrices, with $D_m$ being the dimension after mapping, and $\boldsymbol{K}^{'}$ is the transpose of $\boldsymbol{K}$. Then, the scaled dot-product attention scores $\pmb{\alpha}_{h} \in \mathbb{R}^{N \times N}$ and the hidden representation of load record $\pmb{\psi}_h \in \mathbb{R}^{N \times D_m}$ are computed as
\begin{equation}\label{attention_score_eq}
\begin{split}
&\pmb{\alpha}_{h}= \text{Softmax}\bigg(\frac{ \boldsymbol{Q}_{h}\boldsymbol{K}^{'}_h}{\sqrt{d_{\boldsymbol{K}^{'}_h}}}\bigg);  \pmb{\psi}_h = \pmb{\alpha}_{h} \times\boldsymbol{V}_{h},
\end{split}   
\end{equation}
where $d_{\boldsymbol{K}^{'}_h} = D_m$ is the dimension of the key embedding matrix in head $h$. Thus the final attention is $\pmb{\psi} \in \mathbb{R}^{N \times D} $ after summarizing all $\pmb{\psi}_h$.

\textbf{Feed-forward neural network (FFN)} processes $\pmb{\psi}$, which represents the output after layer normalization across all heads, and passes it through a two-layer multi-layer perceptron with ReLU activation to obtain the final output of the interaction encoder as
\begin{equation}\label{EOS_attention1}
\begin{split}
&\boldsymbol{Z}:= \text{ReLu}(\pmb{\psi}\boldsymbol{\Phi}_1 + \boldsymbol{b}_1)\boldsymbol{\Phi}_2 +\boldsymbol{b}_2,
\\
\end{split}   
\end{equation} where $\boldsymbol{\Phi}_1 \in \mathbb{R}^{D\times D'_m}, \boldsymbol{\Phi}_2 \in \mathbb{R}^{D'_m\times D}$ and $\boldsymbol{b}_1 \in \mathbb{R}^{D'_m}, \boldsymbol{b}_2 \in \mathbb{R}^{D}$ are respectively the learnable weight matrices and bias vectors of the FFN, while $\text{ReLu}(\cdot)$ denotes the ReLU activation function.

As a result, the representation of each sub-sequence is denoted by $\boldsymbol{Z} \in \mathbb{R} ^{N\times D}$, with each vector in the high-dimensional representation encapsulating the information of a sub-sequence.

\subsection{Charging Event Prediction Module}
Building upon the well-learned representation $\boldsymbol{Z}$, which is generated by the previous two modules processing the input load sequence, the charging event prediction module employs a trainable, lightweight fully connected network to obtain the final predicted probability of EV charging state in the near future.

The flattening technique is applied to reshape $\boldsymbol{Z}$, resulting in $\boldsymbol{Z} \in \mathbb{R}^{(N*D) \times 1}$. For each future time step ($m$-th), the probability of EV charging event being ``active'' is computed by
\begin{equation}\label{linear_qe}
    \hat{y}_{m} = \sigma(W'_m\boldsymbol{Z}),
\end{equation}
where $W'_m \in \mathbb{R}^{(N*D) \times 1}$ is the weight matrix for $m$-th minute prediction, and $\sigma(\cdot)$ is the sigmoid activation function.

\subsection{Loss Function}
In the training of DCT-EV, the objective is to minimize the binary cross-entropy loss between the ground-truth labels and the corresponding prediction as follows
\begin{equation}\label{loss}
\mathcal{L}= -\sum_{i}{\sum_m (y_{i,m}\log(\hat{y}_{i,m})) + ((1-y_{i,m})\log(1-\hat{y}_{i,m}))},
\end{equation}
where $y_{i,m}$ is the EV charging event state in the next $m$-th minute for user $i$, and $\hat{y}_{i,m}$ is the corresponding prediction.

\section{Numerical Results and Analysis}\label{sec:results}
In this section, we present the experimental settings, evaluation metrics, and corresponding results and discussions. The simulation experiment employs DCT-EV to predict future EV charging event in minute-interval based on historical meter data. 
As we will show in this section, DCT-EV can accurately predict near-future EV charging event.

\subsection{Experimental Settings and Evaluation}
In our experiment, we preprocess minute-interval residential electricity meter data with EV charging load records in 2018 from Pecan Street \cite{street2016pecan}. 
The dataset provides smart meter data aligned with EV charging profiles, which will be used to construct EV charging label for the experiment. 
After analyzing EV charging profiles, we set the EV charging event label as ``charging" when the EV load is higher than 3 kW, and ``not charging" otherwise. Note that our supervised learning method replies on the EV charging label to learn an effective prediction model.

To set up our experiment, we select homes with EVs and filter out homes without a `charging' label in their records, resulting in 22 households with $8,063,175$ electricity usage records for the experiment. For the selected households, we extract $402,296$ EV charging records in total. In the real-world scenario, meter data is accessible to operators, but the EV charging records are not. Therefore, we use electricity usage as the model input, and EV charging records are merely used for validation. We divide the data into training and testing sets with an 80:20 ratio. The first 80\% of each home's records are for model training, the remaining 20\% are reserved for testing.

The dataset exhibits an imbalance, primarily because EV charging events occur for only a few hours per week, resulting in limited EV charging records. As a classification problem, i.e., predicting future EV charging events, we utilize a set of performance metrics, such as recall, precision, accuracy (ACC), area under the ROC curve (AUC), average precision (AP), and F1 score (F1), along with mean squared error (MSE), to assess the model's overall performance. All these metrics are classical indicators for assessing model performance in machine learning classification problems \cite{hossin2015review}.

In our experiment, the model outputs a probability value representing the likelihood of an active charging event. To compute true positives, true negatives, false positives, and false negatives-- necessary for subsequent F1 score, ACC \footnote{@.5 means threshold is 0.5.} calculations, we adhere to a standard evaluation setup with a threshold set at 0.5. Accordingly, all predictions with a probability exceeding 0.5 are considered indicative of an ``active'' event, while those with a probability less than or equal to 0.5 are interpreted as ``inactive" events.

\subsection{Results and Discussions}

\begin{figure}[t]
    \centering
    \includegraphics[width=.8\linewidth,height=4.3cm]{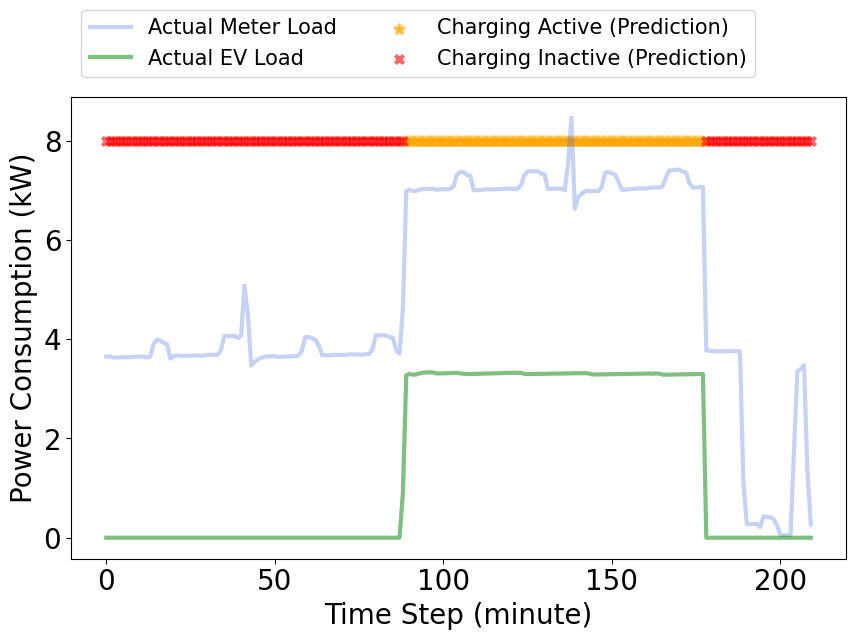} 
    \caption{EV charging event prediction for home $\#$4526.}
    \label{simulation_plot}
\end{figure}

\begin{table}[t]
\centering
\large
\caption{Performance comparisons between our DCT-EV method and baselines, including Random Forest, XGBoost, DNN, and LSTM. (\textcolor{BrickRed}{Red}: maximum and \textcolor{blue}{Blue}: minimum.)}
\label{diff_model:performance}
\resizebox{0.99\linewidth}{!}{\begin{tabular}{c c c c c c c c c c c c}
\toprule
Predication Span & Model Name & F1@.5 & AUC & AP  & ACC@.5 & MSE\\

\multirow{5}{*}{1-10 minutes} &

Random Forest & 22.14\% & 56.50\% & 10.76\%  & 98.55\%  & 0.0144\\
&XGBoost      & 83.43\% & 88.34\% & 71.16\%  & 98.63\%  & 0.0137\\
&DNN          & 83.32\% & 97.55\% & 85.89\%  & 98.45\%  & 0.0122\\
&LSTM         & 83.91\% & 97.73\% & 86.89\%  & 98.74\%  & 0.0103\\

&DCT-EV & \textcolor{BrickRed}{87.35\%} & \textcolor{BrickRed}{97.90\%} & \textcolor{BrickRed}{88.55\%}  & \textcolor{BrickRed}{98.80\%} & \textcolor{blue}{0.0102}\\

\hline

\multirow{3}{*}{1-60 minutes} &
DNN      & 49.74\%  & 88.24\% & 55.20\%  & 96.39\% & 0.0281\\
&LSTM    & 55.94\%  & 88.96\% & 60.06\%  & 96.68\% & 0.0260\\

&DCT-EV & \textcolor{BrickRed}{63.78\%} & \textcolor{BrickRed}{90.07\%} & \textcolor{BrickRed}{61.59\%} & \textcolor{BrickRed}{96.81\%} & \textcolor{blue}{0.0251}\\

\hline
\end{tabular}}
\end{table}

DCT-EV exhibits outstanding performance, as illustrated in Figure~\ref{simulation_plot}, achieving high scores in AUC, AP, F1, ACC, surpassing all compared methods, and establishing a new state-of-the-art benchmark. These results highlight the robustness of the proposed method in predicting EV charging events based on historical smart meter records.

\begin{figure}[t]
    \centering
    \includegraphics[width=0.8\linewidth,height=4cm]{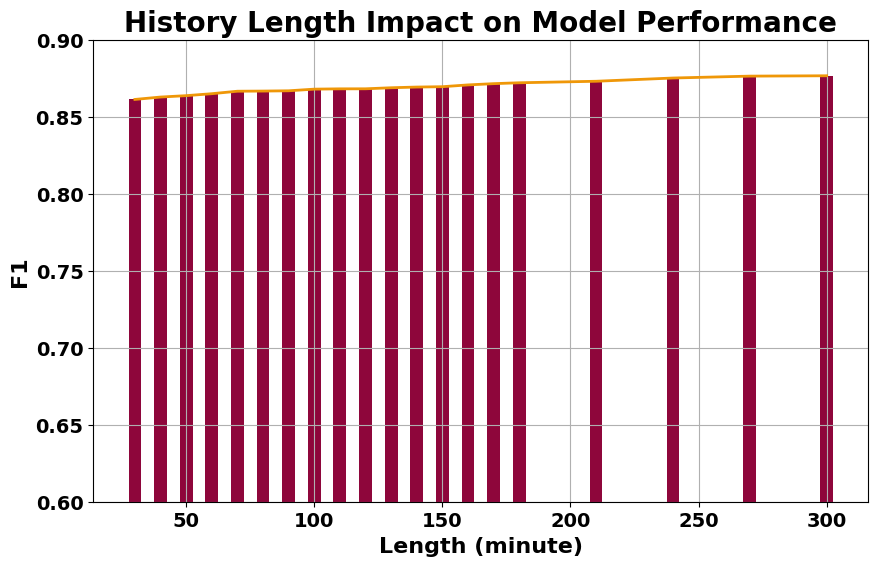}
    \caption{The impact of input history length on the model  performance for predicting EV charging event in the next 60 minutes.}
    \label{hist_plot}
\end{figure}

\subsubsection{Model Comparisons}
To validate the effectiveness of our method, we conduct comprehensive comparative studies with representative baseline models, including Random Forest \cite{breiman2001random}, XGBoost \cite{chen2016xgboost}, Deep Neural Networks (DNN) \cite{murtagh1991multilayer}, and Long Short-Term Memory (LSTM) \cite{hochreiter1997long}. As depicted in Table~\ref{diff_model:performance}, it is evident that DCT-EV outperforms the other models, exhibiting higher F1 scores, AUC, AP, and ACC, as well as lower MSE.

\subsubsection{Impact of Input Length on Model Performance}
The input length can significantly impact model performance, as it influences the amount of information available to the model. As shown in Figure \ref{hist_plot}, the F1 score increases as the length of the size grows, but it plateaus when the size exceeds 180 minutes. While more information contributes to better prediction performance, it also increases computational resources needed. Hence, in this paper, we choose a length size of 180 minutes to tradeoff the performance and computational burden for our subsequent experiments.

\begin{table}[t]
\centering
\large
\caption{The impact of the near-to-distant future prediction on model performance. Threshold @.5. (\textcolor{BrickRed}{Red}: maximum and \textcolor{blue}{Blue}: minimum.)}
\label{diff_span_performance}
\resizebox{1.05\linewidth}{!}{\begin{tabular}{c c c c c c c c c c c c c c }
\toprule

m-th & AUC & AP & F1 & Precision & Recall & ACC & MSE \\
\hline
1-min &\textcolor{BrickRed}{99.63\%}	&\textcolor{BrickRed}{94.87\%}	&\textcolor{BrickRed}{90.88\%}	&\textcolor{BrickRed}{91.36\%}	&\textcolor{BrickRed}{90.40\%}	&\textcolor{BrickRed}{99.15\%}	&\textcolor{blue}{0.0068}\\
2-min &98.96\%	&92.67\%	&89.52\%	&90.39\%	&88.67\%	&99.03\%	&0.0079\\
3-min &98.50\%	&90.97\%	&88.70\%	&89.59\%	&87.83\%	&98.95\%	&0.0086\\
4-min &97.93\%	&89.20\%	&87.73\%	&88.93\%	&86.56\%	&98.86\%	&0.0094\\
5-min &97.30\%	&87.28\%	&86.70\%	&88.06\%	&85.38\%	&98.77\%	&0.0103\\
6-min &96.85\%	&85.55\%	&85.81\%	&87.08\%	&84.58\%	&98.68\%	&0.0110\\
7-min &96.41\%	&83.90\%	&84.79\%	&86.11\%	&83.51\%	&98.59\%	&0.0118\\
8-min &95.95\%	&82.41\%	&83.84\%	&85.25\%	&82.48\%	&98.51\%	&0.0125\\
9-min &95.46\%	&80.71\%	&82.75\%	&84.24\%	&81.31\%	&98.41\%	&0.0133\\
10-min&\textcolor{blue}{94.78\%}	&\textcolor{blue}{78.75\%}	&\textcolor{blue}{81.59\%}	&\textcolor{blue}{83.31\%}	&\textcolor{blue}{79.94\%}	&\textcolor{blue}{98.31\%}	&\textcolor{BrickRed}{0.0142}\\
\hline
\end{tabular}}
\end{table}

\subsubsection{Trade-off Between Model Performance and the Size of Predictive Time Span} 
Our experiments evaluate model performance over two different prediction time spans: next 10 minutes and next 60 minutes. Obviously, all models exhibit better performance in the next 1-10 minutes than 1-60 minutes as illustrated in table \ref{diff_span_performance}. Due to the relatively lower performance of the two machine learning models--XGBoost and random forest in the 1-10 minutes task, we only conduct experiments for the 1-60 minutes ahead prediction for DNN, LSTM, and DCT-EV. All models' performance drop in the 1-60 minutes task; therefore, predicting minute-interval events far into the future is indeed a more challenging task. 

Furthermore, we conduct a series of experiments focused solely on predicting the state in the next m-th minute ($1 \leq m \leq 10$), as shown in Table \ref{diff_span_performance}. The results reveal the same phenomena as discussed above.

\begin{figure}[t]
    \centering
    \includegraphics[width=0.8\linewidth,height=4.5cm]{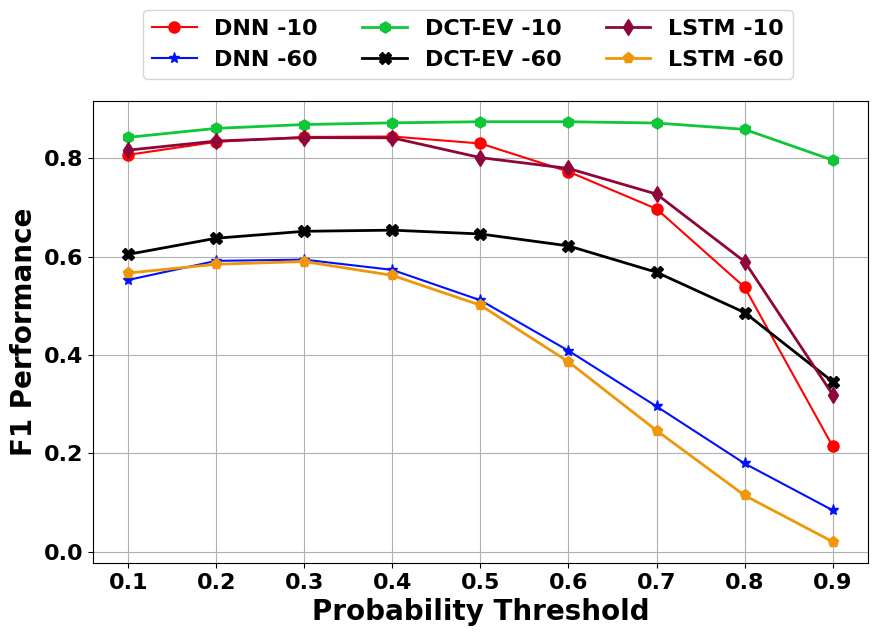}
    \caption{Impact of probability threshold on the performance (F1 score) of DCT-EV, DNN, and LSTM for prediction tasks with 10/60 minutes ahead.}
    \label{threshold_plot}
\end{figure}

\subsubsection{Probability Threshold Selection}
As discussed before, the model's output is the probability of EV charging event. The selection of the probability threshold is crucial to determine whether the prediction is ``active" or ``inactive". We calculate the models' performance (e.g., F1 score) for two different prediction time spans. As depicted in Figure \ref{threshold_plot}, the optimal threshold for DCT-EV falls between 0.6 and 0.8, whereas for DNN and LSTM, it lies in the range of 0.2 to 0.4.
Furthermore, in comparison to DNN and LSTM, DCT-EV exhibits greater robustness to threshold selection effects.

\section{Conclusion} \label{sec:conclu}
Accurate prediction of EV charging events is paramount for EV energy management and grid operations. Existing EV charging prediction methods, tailored to charging stations, overlooked the need for at-home EV charging predictions, especially when historical EV charging profile is not available. To address this problem, we introduced DCT-EV, a deep learning model leveraging parallel processing and self-attention mechanisms, to predict minute-by-minute future EV charging events with remarkable accuracy, exceeding 96\%. Importantly, DCT-EV relies solely on historical smart meter data instead of EV charging profile, making it a valuable tool for grid management in the era of high EV adoption.

\bibliographystyle{unsrt}
\bibliography{ref.bib}

\end{document}